\documentclass{article}

\usepackage[preprint]{corl_2026} 

\usepackage{amsmath,amssymb,amsthm,mathtools}
\usepackage{booktabs}
\usepackage{multirow}
\usepackage{array}
\usepackage{algorithm}

\usepackage{algpseudocode}
\usepackage{caption}

\newcommand{\oracle}{\mathcal{O}_{\mathrm{s}}}
\newcommand{\safe}{\mathcal{C}}

\newtheorem{proposition}{Proposition}

\title{Learning Safe Flow-Matching Robot Policies from Binary Feasibility Oracles}

\author{
  Jianming Ma\\
  Shanghai Jiao Tong University\\
  \And
  Rongjun Jin \\
  Shanghai Jiao Tong University \\
  \AND
  Yizhi Chen \\
  Tongji University \\
  \And
  Yue Gao \\
  Shanghai Jiao Tong University \\
}

\hypersetup{
  pdftitle={Learning Safe Flow-Matching Robot Policies from Binary Feasibility Oracles},
  pdfauthor={Jianming Ma, Rongjun Jin, Yizhi Chen, Yue Gao},
  pdfsubject={Preprint}
}

\begin{document}
\maketitle


\begin{abstract}
Robot foundation models increasingly rely on flow-based action generation for general-purpose manipulation, yet the generated actions are not guaranteed to satisfy physical constraints such as joint limits and collision avoidance. This problem is especially challenging when the feasible set cannot be described analytically and is accessible only through black-box feasibility oracles. We study this practical \emph{oracle-only} setting and propose an oracle-guided differentiable safeguard for safe flow-matching robot policies. The safeguard uses binary feasibility queries to correct unsafe updates in the forward pass, while extracting local boundary information to provide constraint-aware learning signals in the backward pass. Combined with discrete-time flow matching, the same safeguard is used during training and inference, enabling forward--backward aligned policy learning without explicit or differentiable constraint models. Experiments on obstacle-aware robotic manipulation demonstrate 100\% constraint satisfaction while preserving task performance.

\end{abstract}


\section{Introduction}
\label{sec:introduction}

Robot foundation models, including vision-language-action (VLA) models \citep{intelligence2025pi_,bjorck2025gr00t} and world-action models (WAMs) \citep{ye2026world,yuan2026fast}, increasingly combine pretrained vision-language or video-generation backbones with flow-based action generation for general-purpose robot control. However, strong task generalization does not guarantee safe physical interaction: generated actions must also satisfy constraints such as joint limits and collision avoidance. Enforcing such scene-dependent constraints remains a key challenge for deploying flow-based robot policies.

Existing methods either correct generated actions only at inference time or incorporate constraint enforcement into training. Inference-time correction, e.g., through control barrier functions \citep{ames2017control,yang2026safeflowmatcher,huang2025sad,hu2025vlsa}, or trajectory-wise constraint guidance \citep{utkarsh2026physics,cheng2025gradient}, can be applied to pretrained policies but introduces a train--test mismatch. Training-aware approaches reduce this mismatch \citep{ma2026actsafeguard,christopher2026constraint,li2026gauge}, but typically require explicit constraint functions, differentiable constraint models, or known feasible-set geometry. Such information is often difficult to obtain in robotics. For example, in robotic arm obstacle avoidance, a collision checker can readily determine whether a candidate configuration is feasible, while an analytical description of the corresponding collision-free configuration space may be unavailable.

This motivates an \emph{oracle-only} setting, where safety is accessible solely through a binary feasibility oracle. We ask: \emph{how can a flow-matching robot policy enforce constraints and learn from them when only binary feasibility queries are available?} To address this problem, we propose an \emph{oracle-guided differentiable safeguard}. In the forward pass, the safeguard uses oracle queries to restrict each predicted update to the feasible region. In the backward pass, it extracts local boundary information from nearby binary queries and uses it to construct a boundary-aware surrogate gradient. Thus, the same oracle supports both feasibility enforcement and constraint-aware policy learning, without requiring an explicit or differentiable constraint model.

We integrate the safeguard with discrete-time flow matching, allowing the same correction mechanism to be used during both training and inference. This yields a forward--backward aligned learning framework in which the policy is optimized under the safety intervention encountered at deployment.

Our contributions are summarized as follows:
\begin{itemize}
    \item We formulate safe flow-matching robot policy learning under an \emph{oracle-only} constraint interface, removing the need for explicit or differentiable constraint models.
    
    \item We introduce an oracle-guided differentiable safeguard that combines oracle-based forward feasibility correction with boundary-aware surrogate gradients, enabling forward--backward aligned policy learning.
    
    \item We validate the proposed approach on an obstacle-aware robotic manipulation task, demonstrating strong constraint satisfaction while maintaining task performance.
\end{itemize}

\section{Preliminaries and Problem Formulation}
\label{sec:prelim_problem}

Consider a robot manipulation task at control step $t$, where $o_t$ denotes
the current multimodal observation and language instruction. The policy predicts an $H$-step future joint-position trajectory
$
    A_t
    =
    [q_t^{(1)},\ldots,q_t^{(H)}]
    \in\mathbb{R}^{H\times d},
    \label{eq:action_chunk}
$
where $q_t^{(h)}\in\mathbb{R}^d$ denotes the predicted joint configuration
at the $h$-th future horizon. The task-dependent safety constraints are accessible only through a binary
feasibility oracle
\begin{equation}
    \oracle(q;\xi_t)\in\{0,1\},
    \qquad
    \safe(\xi_t)
    =
    \{q\in\mathbb{R}^d:\oracle(q;\xi_t)=1\},
    \label{eq:oracle}
\end{equation}
where $\xi_t$ denotes the task context, and $\safe(\xi_t)$ denotes the feasible region. The oracle need not be
differentiable and does not provide an analytical description of the
feasible set. Our goal is to ensure that every configuration in the
predicted action chunk is feasible:
$
    q_t^{(h)}\in\safe(\xi_t),
    \;
    \forall h\in\{1,\ldots,H\}.
    \label{eq:safety_objective}
$

\paragraph{Discrete-time flow matching.}
We generate the entire action chunk through a discrete-time  flow. Let
$
    X_k
    =
    [x_k^{(1)},\ldots,x_k^{(H)}]
    \in\mathbb{R}^{H\times d},
    \;
    k=0,\ldots,T,
    \label{eq:flow_chunk}
$
denote the action-chunk sample at flow step $k$. Starting from
$X_0\sim p_0(\cdot\mid o_t)$, the action expert predicts the finite update
\begin{equation}
    D_k
    =
    \Delta_\theta(X_k,k,o_t)
    =
    [d_k^{(1)},\ldots,d_k^{(H)}],
    \qquad
    X_{k+1}=X_k+D_k.
    \label{eq:discrete_flow}
\end{equation}

Given a target action chunk $A_t$, we set $X_T=A_t$ and use the linear
interpolation~\citep{liu2023rectified}
$
    X_k
    =
    \left(1-k/T\right)X_0
    +
    k/T \cdot A_t.
    \label{eq:linear_path}
$
The corresponding discrete flow-matching objective is
\begin{equation}
    \mathcal{L}_{\mathrm{DFM}}
    =
    \mathbb{E}
    \left[
        \left\|
        \Delta_\theta(X_k,k,o_t)
        -
        (A_t-X_0)/T
        \right\|_F^2
    \right],
    \label{eq:dfm_loss}
\end{equation}
where $\|\cdot\|_F$ denotes the Frobenius norm over the entire action
chunk.

This finite-update formulation provides a natural interface for enforcing safety during both training and inference. Starting from a feasible initial sample, feasibility can be preserved recursively by safeguarding each discrete update, thereby ensuring that the final generated sample remains in the feasible set.

\section{Oracle-Guided Differentiable Safeguard}
\label{sec:safeguard_layer}

We introduce a safeguard that uses the binary oracle for both forward
feasibility correction and backward boundary-aware learning. Since the
safety oracle is defined over individual joint configurations, the
safeguard is applied independently to each horizon element
$(x_k^{(h)},d_k^{(h)})$. For clarity, we omit the horizon index $h$ below
and write $x_k$ and $d_k$ for a single horizon element.

\paragraph{Forward: Oracle-Guided Safety Correction}
\label{sec:safeguard_forward}

Given a feasible point $x_k$ and its nominal update $d_k$, we
reparameterize the transition as
$
    x_{k+1}
    =
    x_k+\alpha_k d_k,
    \;
    \alpha_k\in[0,1].
    \label{eq:safe_reparameterization}
$
Rather than reconstructing the feasible set, we search only along the ray
$
    r_k(\alpha)
    =
    x_k+\alpha d_k,
    \;
    \alpha\in[0,1],
$
for the first exit from the feasible region. The ideal first-exit factor is
\begin{equation}
    \alpha_k^*
    =
    \sup
    \left\{
        \alpha\in[0,1]:
        \oracle(r_k(\tau);\xi_t)=1,\;
        \forall\tau\in[0,\alpha]
    \right\}.
    \label{eq:first_exit_scaling}
\end{equation}

Since only binary queries are available, we approximate $\alpha_k^*$ using
an oracle-based ray search, e.g., a coarse scan followed by local
bisection:
$
    \hat{\alpha}_k
    =
    \operatorname{OracleRaySearch}
    (\oracle,x_k,d_k;\xi_t).
    \label{eq:oracle_ray_search}
$
The safeguarded update is then
$
    \Delta_k^{\mathrm{safe}}
    =
    \hat{\alpha}_k d_k,
    \;
    x_{k+1}
    =
    x_k+\Delta_k^{\mathrm{safe}}.
    \label{eq:safeguarded_forward}
$

\paragraph{Backward: Oracle-Guided Boundary Gradient}
\label{sec:safeguard_backward}

The oracle-based forward correction is non-differentiable. To expose the
encountered constraint boundary to the action expert during training, we
construct a boundary-aware surrogate Jacobian.

Assuming a locally smooth boundary with outward normal $n_k$, when the update is truncated at the boundary, its Jacobian with respect to the nominal update is
\begin{equation}
    \frac{\partial \Delta_k^{\mathrm{safe}}}{\partial d_k}
    =
    \alpha_k^*
    \left(
        I-\frac{d_k n_k^\top}{n_k^\top d_k}
    \right).
    \label{eq:boundary_aware_jacobian}
\end{equation}
For an untruncated update, the Jacobian reduces to $I$.

Since $n_k$ is unavailable under the oracle-only setting, we estimate it
from symmetric binary feasibility queries near the detected boundary.
Let $z_c$ denote a point near the boundary and
$\{u_i\}_{i=1}^{K}$ a set of unit probing directions. For each direction,
we query
$
    z_i^\pm
    =
    z_c\pm\rho u_i,
$
and define the directional vote
$
    s_i
    =
    \oracle(z_i^-;\xi_t)
    -
    \oracle(z_i^+;\xi_t).
    \label{eq:directional_vote}
$
Aggregating these votes gives the approximate outward normal
$
    \hat n_k
    =
    (
        \sum_{i=1}^{K}s_i u_i
    )/(
        \left\|
        \sum_{i=1}^{K}s_i u_i
        \right\|_2+\epsilon_n
    ),
    \;
    \hat n_k^\top d_k\geq0.
    \label{eq:normal_estimate}
$
Substituting $\hat n_k$ into
Eq.~\eqref{eq:boundary_aware_jacobian} yields the practical surrogate
Jacobian. The same oracle therefore serves two complementary roles:
forward feasibility enforcement and backward estimation of local boundary
geometry.

\begin{figure*}[t]
    \centering

    \begin{minipage}[t]{0.485\textwidth}
        \vspace{0pt}
        \centering
        \includegraphics[
            width=\linewidth,
            keepaspectratio
        ]{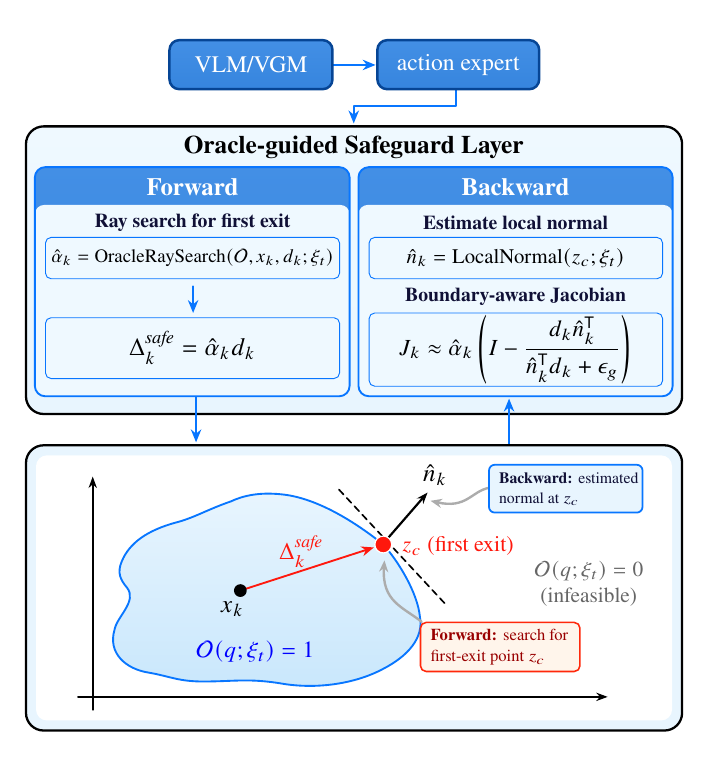}
    \end{minipage}
    \hfill
    \begin{minipage}[t]{0.485\textwidth}
    \vspace{0pt}
    \raggedright
    \captionsetup{
        type=algorithm,
        justification=raggedright,
        singlelinecheck=false,
        font=small
    }

    \hrule
    \vspace{2pt}

    \captionof{algorithm}{Safe Discrete-Time Flow-Matching Training}
    \label{alg:safe_dfm_training}

    \vspace{-2pt}
    \hrule
    \vspace{4pt}

    \small
    \begin{algorithmic}[1]
    \Require Dataset $\mathcal{D}$, action expert $\Delta_\theta$, oracle $\oracle$
    \Ensure Trained parameters $\theta$

    \Repeat
        \State Sample $(o_t,\xi_t,A_t)\sim\mathcal{D}$
        \State Sample feasible $X_0\sim p_0(\cdot\mid o_t)$
        \State Construct reference path
        $\{X_k^\star\}_{k=0}^{T}$ with $X_T^\star=A_t$
        \State Sample $k\sim\mathcal{U}\{0,\ldots,T-1\}$

        \State $D_k^\star
            \gets
            X_{k+1}^\star-X_k^\star$

        \State $D_k
            \gets
            \Delta_\theta(X_k^\star,k,o_t)$

        \State $(D_k^{\mathrm{safe}},\mathcal{J}_k)
            \gets
            \Call{HorizonWiseSafeguard}
            {X_k^\star,D_k,\xi_t,\oracle}$

        \State $\mathcal{L}
            \gets
            \|D_k^{\mathrm{safe}}-D_k^\star\|_F^2$

        \State Update $\theta$ using
        $\nabla_\theta\mathcal{L}$ through $\mathcal{J}_k$
    \Until{convergence}

    \State \Return $\theta$
    \end{algorithmic}

    \vspace{4pt}
    \hrule
    \end{minipage}

    \caption{
    \textbf{Left:} Overview of oracle-guided differentiable safeguard.
    \textbf{Right:} safe discrete-time flow-matching training using the same safeguard.
    }
    \label{fig:method_overview}

\end{figure*}

\section{Experimental Results}
\label{sec:experiments}

\begin{figure}
    \centering
    \includegraphics[width=\linewidth]{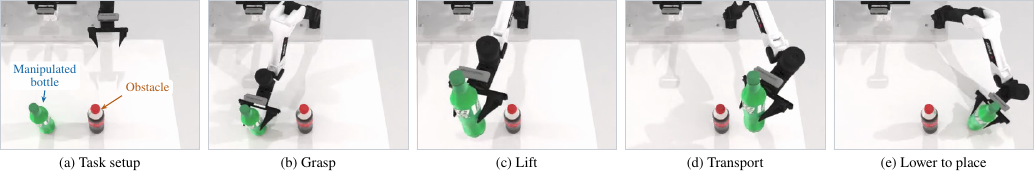}
    \caption{Snapshots of a rollout using $\pi_{0.5}$ + safeguard, subject to collision-avoidance constraints and joint-position limits.}
    \label{fig:task-rollout}
\end{figure}

We design a single-arm pick-and-place task in RoboTwin2.0 \citep{chen2025robotwin}, as illustrated in Fig.~\ref{fig:task-rollout}. The robot moves a bottle from left to right while avoiding a stationary bottle obstacle and the tabletop, subject to joint-position limits. The feasibility oracle is implemented using privileged information from the simulator.

We use absolute joint-position actions with horizon $H=20$ and fine-tune pretrained $\pi_{0.5}$ models using LoRA~\citep{hu2021lora} on 150 expert obstacle-avoidance trajectories for 30K steps on a single NVIDIA H100 GPU.

\paragraph{Baselines.}
We compare the following variants: (1) $\pi_{0.5}$ baseline: the original model without our safeguard. \textit{$\pi_{0.5}$ (gaussian init)} uses a standard Gaussian initial distribution, while \textit{$\pi_{0.5}$ (local init)} samples initial points around the current observed joint configuration. (2) \textit{$\pi_{0.5}$ + safeguard}: the baseline augmented with our oracle-guided differentiable safeguard. (3) \textit{$\pi_{0.5}$ + safeguard + stop grad}: an ablation of our method that removes the boundary-aware backward gradient estimation. Our method requires the initial sample to lie in the feasible region. We therefore sample around the current observed joint configuration and use the oracle to filter out infeasible samples.

\paragraph{Evaluation.}
We evaluate each method for 30 episodes under the clean setting. We report episode success rate, step safety rate, motion smoothness measured by LDLJ~\citep{balasubramanian2011robust}, and the average inference time per action chunk. As shown in Table~\ref{tab:main_results}, $\pi_{0.5}$ + safeguard achieves a $100\%$ step safety rate while maintaining a success rate close to the unconstrained baseline. It also improves LDLJ from $-22.99$ to $-18.35$, indicating that the safeguard improves safety without severely sacrificing task performance and can also produce smoother motions. The $\pi_{0.5}$ (local init) baseline shows that these gains do not come from the local initialization alone, as it yields much lower success and worse smoothness than $\pi_{0.5}$ + safeguard. Moreover, the stop-gradient ablation degrades both success and smoothness, suggesting that propagating boundary information back to the policy helps it better adapt to the constrained feasible region.

\begin{table*}[t]
    \centering
    \caption{Comparison of task success, safety, motion smoothness, and inference efficiency. Best results are shown in bold.}
    \label{tab:main_results}
    \small
    \renewcommand{\arraystretch}{0.9}
    \setlength{\tabcolsep}{8pt}
    \begin{tabular*}{\textwidth}{
        @{\extracolsep{\fill}}lcccc@{}
    }
        \toprule
        Method & Success Rate $\uparrow$
               & Step Safety Rate $\uparrow$
               & LDLJ $\uparrow$ & Inference Time (s) $\downarrow$\\
        \midrule
        $\pi_{0.5}$ (gaussian init) & \textbf{13/30} & 78.25\% & -22.99 & \textbf{0.071}\\
        $\pi_{0.5}$ (local init) & 0/30 & 94.12\% & -22.55 & 0.076\\
        $\pi_{0.5}$ + safeguard + stop grad & 8/30 & \textbf{100}\% & -19.32 & 1.808\\
        $\pi_{0.5}$ + safeguard & 11/30 & \textbf{100}\% & \textbf{-18.35} & 1.345\\
        \bottomrule
    \end{tabular*}
\end{table*}

\section{Conclusion}
\label{sec:conclusion}

We propose an oracle-guided differentiable safeguard for safe flow-matching robot policies using only binary feasibility feedback. The oracle enforces feasibility in the forward pass and provides local boundary information for backward learning, allowing the same safeguard to be used during training and inference without explicit or differentiable constraint models. Experiments on obstacle-aware manipulation show improved safety while preserving task performance and motion quality.

A current limitation is that the discrete flow still relies on simple interpolation paths, which may become unreliable in highly non-convex feasible regions. Future work will explore constraint-aware flow paths and evaluate the method across more diverse tasks, robots, and real-world constraints.


\clearpage


\bibliography{references}  

\clearpage
\appendix
\numberwithin{equation}{section}

\section{Derivation of the Boundary-Aware Backward Pass}
\label{app:derivation}

Following the notation in Section~\ref{sec:safeguard_layer}, consider a feasible point $x_k$ whose update $d_k$ is truncated at $x_k+\alpha_k^*d_k$. To derive Eq.~\eqref{eq:boundary_aware_jacobian}, hold $x_k$ fixed and perturb $d_k$. The resulting displacement of the boundary intersection lies in the local tangent plane and is therefore orthogonal to the outward normal $n_k$. This gives
\begin{equation}
    n_k^\top\left(\alpha_k^*\,\mathrm{d}d_k+d_k\,\mathrm{d}\alpha_k^*\right)=0,
    \qquad
    \nabla_{d_k}\alpha_k^*
    =-\frac{\alpha_k^*n_k}{n_k^\top d_k},
    \label{eq:app_alpha_differential}
\end{equation}
where $n_k^\top d_k\neq0$ at a transverse intersection. Applying the product rule to the ideal corrected update $\Delta_k^{\mathrm{safe}}=\alpha_k^*d_k$ then yields
\begin{equation}
    \frac{\partial\Delta_k^{\mathrm{safe}}}{\partial d_k}
    =\alpha_k^*I+d_k\left(\nabla_{d_k}\alpha_k^*\right)^\top
    =\alpha_k^*\left(I-\frac{d_kn_k^\top}{n_k^\top d_k}\right),
    \label{eq:app_update_jacobian}
\end{equation}
which recovers Eq.~\eqref{eq:boundary_aware_jacobian}. For an untruncated update, $\alpha_k^*=1$ is locally constant, so the Jacobian reduces to $I$. In practice, $\hat\alpha_k$ and $\hat n_k$ provide oracle-based approximations to the intersection factor and its normal, as described next.

\section{Practical Implementation}
\label{app:implementation}

\paragraph{Forward: coarse scanning and local bisection.}
Given $x_k$ and $d_k$, we first query the oracle at uniformly spaced points along $r_k(\alpha)=x_k+\alpha d_k$, with $\alpha\in[0,1]$. If all sampled points are feasible, the full update is accepted. Otherwise, the first sampled infeasible point and its preceding feasible point define a local search interval. Then we refine this interval by bisection. The final feasible endpoint approximates the first-exit point. 

\paragraph{Backward: local boundary estimation.}
The final search interval also provides the neighborhood for estimating $n_k$. We place $z_c$ at its midpoint and query symmetric pairs $z_i^\pm=z_c\pm\rho u_i$, using coordinate axes and random unit probing directions $\{u_i\}_{i=1}^{K}$. When the two labels differ, the vote $s_i=\oracle(z_i^-;\xi_t)-\oracle(z_i^+;\xi_t)$ gives a signed contribution $s_i u_i$ toward the infeasible side; matching labels contribute zero. Aggregating and normalizing these contributions gives $\hat n_k$, with its sign chosen so that $\hat n_k^\top d_k\geq0$, as in the main text. This use of binary labels to estimate boundary directions is related to decision-based methods such as HopSkipJumpAttack~\citep{chen2020hopskipjump}. Here, the estimate supplies the local boundary information for the surrogate backward pass. The queries are batched across horizon elements, and normal estimation is needed only during training.

\section{Experimental Details}
\label{app:experiment_details}

\paragraph{Expert data collection.}
We collect demonstrations with a collision-aware MPlib planner. Inverse kinematics first provides candidate grasp configurations, after which RRT plans joint-space motions to the pre-grasp and pre-place poses. Local Cartesian planning completes the short approach and placement motions. Once the grasp is physically verified, the bottle geometry is attached to the end effector in the planning scene so that subsequent transport accounts for the carried object. The planner also uses a 4 cm inflated obstacle envelope to leave clearance along the planned motion. We retain successful, safety-checked episodes and replay their stored joint paths to record synchronized observations, actions, and privileged scene information, yielding the 150 demonstrations used in Table~\ref{tab:main_results}.

\paragraph{Oracle implementation.}
The recorded scene information supports a batched JAX feasibility oracle that combines joint-limit checks with collision checks. For each queried configuration, forward kinematics transforms sphere approximations of the left-arm links into world coordinates, where they are checked against the scene geometry. During transport, the carried bottle is placed using the candidate end-effector pose and the observed object-to-end-effector transform, allowing its collisions to be checked as well. The safeguard accesses the resulting binary feasibility labels without differentiating through the checker.

\paragraph{Feasible initial distribution.}
We construct $p_0(\cdot\mid o_t)$ by randomly sampling joint configurations in a local neighborhood of the currently observed joint state and retaining candidates that satisfy the oracle constraints. An accepted configuration is repeated across the horizon to form $X_0$. The same initialization procedure is used during training and inference, providing a feasible starting point for the safeguarded flow.

\paragraph{Training parameters.}
All variants are fine-tuned with LoRA for 30K updates on a single H100 GPU, using a batch size of 32. We use AdamW with $(\beta_1,\beta_2)=(0.9,0.95)$, $\epsilon=10^{-8}$, weight decay $10^{-10}$, and gradient clipping at 1.0. The learning rate increases to $2.5\times10^{-5}$ over 1K warmup steps and then follows a cosine schedule, with the decay step set to 20K and a final rate of $2.5\times10^{-6}$. An exponential moving average of the parameters is maintained with decay 0.99.

\section{Discussion and Limitations}
\label{app:discussion}

\subsection{Why Maintain Feasibility at Every Generation Step?}

Intermediate flow states are not executed by the robot, so their feasibility is an algorithmic choice rather than a physical execution requirement. We make this choice to construct a generation process that remains within the feasible set: starting from a feasible sample, each safeguarded update preserves feasibility, which in turn ensures that the final output satisfies the constraints. This shares the feasible-iterate motivation of interior-point methods, although our method uses oracle-based correction rather than a barrier objective. Maintaining generation trajectories within a constrained domain is also central to reflected flow matching~\citep{xie2024reflected}, which uses boundary reflection. The resulting discrete invariance of our method is stated below.

\begin{proposition}[Fixed-context feasibility preservation]
\label{prop:app_invariance}
For a fixed context $\xi_t$, suppose $X_0\in\safe(\xi_t)^H$ and each horizon-wise update returns an oracle-verified feasible configuration or retains its feasible input. Then $X_k\in\safe(\xi_t)^H$ for every $k=0,\ldots,T$.
\end{proposition}
The claim follows by induction, since the forward search returns a checked feasible point at each step. Applying the same correction during training also exposes the policy to the interventions it encounters during inference. Together, these choices align policy learning with feasible generation and make final configuration feasibility a property of the sampling procedure, without requiring a separate terminal projection.

\subsection{What Are the Main Limitations?}

\paragraph{Feasible data, initialization, and local geometry.}
The formulation assumes that the target action distribution lies within the feasible set and that feasible initial samples can be obtained. With linear reference paths, these endpoint conditions must be accompanied by local feasible connectivity: a sufficient condition is that each source--target pair lies within a common convex feasible neighborhood. In a non-convex region, a straight reference path may instead cross an obstacle even when both endpoints are feasible, creating supervision that the safeguarded update cannot follow. The forward correction also preserves the predicted direction, so it cannot independently select a route around the obstacle. These limitations motivate reference paths that better capture feasible task geometry.

\paragraph{Oracle fidelity and computational cost.}
The practical correction depends on finite scans and local probes, whose resolution limits how accurately they capture complex boundaries. Its safety scope is therefore determined by the oracle and the configurations actually checked. Increasing the query budget can improve boundary resolution, but also increases computation. In Table~\ref{tab:main_results}, the full method requires 1.345 s per chunk compared with 0.071 s for the Gaussian baseline, approximately $18.9\times$ as long. Reducing this overhead while retaining reliable feasibility checks is necessary for more responsive manipulation.

\subsection{Future Directions}

A natural next step is to replace straight reference paths with nonlinear paths that remain within the feasible region. Riemannian flow matching~\citep{chen2024geometry} illustrates how known manifold geometry can inform the construction of target vector fields; in our oracle-only setting, feasible expert trajectories offer a practical source of path information. The reference can follow the demonstrated route around obstacles rather than directly interpolate between the initial sample and each target configuration. With feasible local connections, this could extend the approach to non-convex constraint regions while retaining the same oracle-guided safeguard. The remaining challenge is to connect randomly sampled initial configurations to these paths and discretize them without introducing infeasible shortcuts.

A complementary direction is to reduce the cost of oracle queries during generation. Adaptive coarse scanning and selective local refinement could concentrate queries near detected boundaries, while more efficient batched collision checking could reduce their latency. Evaluating these strategies jointly with task success and constraint satisfaction would help identify a practical balance between computational cost and safety for online manipulation.

\end{document}